%% file: example_paper.tex
\documentclass{article}

\usepackage{times}
\usepackage{graphicx} 

\usepackage{natbib}

\usepackage{algorithm}
\usepackage{algorithmic}
\usepackage{graphicx}
\usepackage{amsfonts}
\usepackage{amsmath}
\usepackage{amsthm}
\usepackage[percent]{overpic}
\usepackage{color}
\usepackage{caption}
\usepackage{subcaption}
\usepackage{xspace}
\usepackage{verbatim}

\newcommand{\pbim}{\textsc{PSIM}\xspace}
\newcommand{\PBIM}{\textsc{Predictive State Inference Machine}\xspace}

\newcommand{\pbims}{\textsc{PSIM}s\xspace}
\newcommand{\PBIMs}{\textsc{Predictive State Inference Machines}\xspace}

\newtheorem{theorem}{Theorem}[section]

\newtheorem{proposition}[theorem]{Proposition}
\newtheorem{corollary}[theorem]{Corollary}

\usepackage[accepted]{icml2016}

\icmltitlerunning{Learning to Filter with Predictive State Inference Machines}

\begin{document} 

\twocolumn[
\icmltitle{Learning to Filter with Predictive State Inference Machines}

\icmlauthor{Wen Sun$^{\dagger}$}{wensun@cs.cmu.edu}
\icmlauthor{Arun Venkatraman$^{\dagger}$}{arunvenk@cs.cmu.edu}
\icmlauthor{Byron Boots$^*$}{bboots@cc.gatech.edu}
\icmlauthor{J. Andrew Bagnell$^{\dagger}$}{dbagnell@ri.cmu.edu}
\icmladdress{$^\dagger$Robotics Institute, Carnegie Mellon University, USA \\
$^*$College of Computing, Georgia Institute of Technology, USA 
}
\icmlkeywords{boring formatting information, machine learning, ICML}

\vskip 0.3in
]

\begin{abstract}
Latent state space models are a fundamental and widely used tool for modeling dynamical systems. 
 However, they are difficult to learn from data and learned models often lack performance guarantees on inference tasks such as filtering and prediction. In this work, we present the \PBIM (\pbim), a data-driven method that considers the inference procedure on a dynamical system as a composition of predictors. %
 The key idea is that rather than first learning a latent state space model, and then using the learned model for inference, 
 \pbim directly learns predictors for inference in predictive state space. 
 We provide theoretical guarantees for inference, in both realizable and agnostic settings, and showcase practical performance on a variety of simulated and real world robotics benchmarks.
\end{abstract}

\section{Introduction}
\label{sec:intro}
Data driven approaches to modeling dynamical systems is important in applications ranging from time series forecasting for market predictions to filtering in robotic systems. The classic generative approach is to assume that each observation is correlated to the value of a latent state and then model the dynamical system as a graphical model, or latent state space model, such as a \emph{Hidden Markov Model} (HMM). To learn the parameters of the model from observed data, Maximum Likelihood Estimation (MLE) based methods attempt to maximize the likelihood of the observations with respect to the parameters. This approach has proven to be highly successful in some applications~\citep{Coates2008, roweis1999unifying}, but has at least two shortcomings. First, it may be difficult to find an appropriate parametrization for the latent states. If the model is parametrized incorrectly, the learned model may exhibit poor performance on inference tasks such as Bayesian filtering or predicting multiple time steps into the future. Second, learning a latent state space model is difficult. The MLE objective is non-convex 
and finding the globally optimal solution is often computationally infeasible. Instead, algorithms such as Expectation-Maximization (EM) are used to compute locally optimal solutions. Although the maximizer of the likelihood objective can promise good performance guarantees when it is used for inference, the locally optimal solutions returned by EM typically do not have any performance guarantees.

\emph{Spectral Learning} methods are a popular alternative to MLE for learning models of dynamical systems~\citep{Boots2012_Thesis,Boots2011_IJRR,Hsu2009_COLT,NIPS2015_5673}. This family of algorithms provides theoretical guarantees on discovering the global optimum for the model parameters under the assumptions of infinite training data and realizability. However, in the non-realizable setting --- i.e. model mismatch (e.g., using learned parameters of a Linear Dynamical System (LDS) model for a non-linear dynamical system) --- these algorithms lose any performance guarantees on using the learned model for filtering or other inference tasks. For example, \citet{kulesza2014low} shows when the model rank is lower than the rank of the underlying dynamical system, the inference performance of the learned model may be arbitrarily bad.

Both EM and spectral learning suffer from limited theoretical guarantees: 
from model mismatch for spectral methods, and from computational hardness for finding the global optimality of non-convex objectives for MLE-based methods. 
In scenarios where our ultimate goal is to infer some quantity from observed data, 
a natural solution is to skip the step of learning a model, and instead directly optimize the inference procedure. 
%
%
%
%
Toward this end, we generalize the \textit{supervised message-passing Inference Machine} approach of~\citet{ross2011_CVPR, ramakrishna2014pose, lin2015deeply}. Inference machines do not parametrize the graphical model (e.g., design of potential functions) and instead directly train predictors that use incoming messages and local features to predict outgoing messages via black-box supervised learning algorithms. 
By combining the model and inference procedure into a single object --- an \emph{Inference Machine} --- we directly optimize the end-to-end quality of inference. 
This unified perspective of learning and inference enables stronger theoretical guarantees on the inference procedure: the ultimate task that we care about.      

   
One of the principal limitations of inference machines is that they require supervision. If we only have access to observations during training, then there is no obvious way to apply the inference machine framework to graphical models with latent states.  
To generalize Inference Machines to dynamical systems with latent states, we leverage ideas from \emph{Predictive State Representations} (PSRs)~\citep{Littman01,Singh2004_UAI, Boots2011_IJRR, NIPS2015_5673}. 
In contrast to latent variable representations of dynamical systems, which represent the belief state as a probability distribution over the {unobserved} state space of the model, PSRs instead maintain an \emph{equivalent} belief over sufficient features of future observations. 

We propose \PBIMs (\pbims), an algorithm that treats the inference procedure (filtering) on a dynamical system as a composition of predictors. Our procedure takes the current predictive state and the latest observation from the dynamical system as inputs and outputs the next predictive state (Fig.~\ref{fig:ps_filter}). Since we have access to the observations at training, this immediately brings the supervision back to our learning problem --- we quantify the loss of the predictor by measuring the likelihood that the actual future observations are generated from the predictive state computed by the learner. \pbim allows us to treat filtering as a general supervised learning problem handed-off to a black box learner of our choosing. The complexity of the learner naturally controls the trade-off between computational complexity and prediction accuracy. We provide two algorithms to train a \pbim. The first algorithm learns a sequence of non-stationary filters which are provably consistent in the realizable case. The second algorithm is more data efficient and learns a stationary filter which has reduction-style performance guarantees. 

The three main contributions of our work are: (1) we provide a reduction of unsupervised learning of latent state space models to the supervised learning setting by leveraging PSRs; (2) our algorithm, \pbim, directly minimizes error on the inference task---closed loop filtering; (3) \pbim works for general \emph{non-linear} latent state space models and guarantees filtering performance even in agnostic setting.

\section{Related Work}
In addition to the MLE-based approaches and the spectral learning approaches mentioned in Sec.~\ref{sec:intro}, there are several supervised learning approaches related to our work. Data as Demonstrator (DaD) \citep{venkatraman2015improving} applies the Inference Machine idea to fully observable Markov chains, and directly optimizes the \emph{open-loop} forward prediction accuracy. 
In contrast, we aim to design an \emph{unsupervised} learning algorithm for \emph{latent} state space models (e.g., HMMs and LDSs) to improve the accuracy of \emph{closed loop} prediction--Bayesian filtering. It is unclear how to apply DaD to learning a Bayesian filter. Autoregressive models \cite{wei1994time} on $k$-th order fully observable Markov chains (AR-$k$) use the most recent $k$ observations to predict the next observation. The AR model is not suitable for latent state space models since the beliefs of latent states are conditioned on the \emph{entire} history. Learning mappings from entire history to next observations is unreasonable and one may need to use a large $k$ in practice. A large $k$, however, increases the difficulty of the learning problem (i.e., requires large computational and samples complexity). 

In summary, our work is conceptually different from DaD and AR models in that we focus on unsupervised learning of latent state space models. Instead of simply predicting next observation, we focus on predictive state---a distribution of future observations, as an alternative representation of the beliefs of latent states. 


\section{Preliminaries}
We consider uncontrolled discrete-time time-invariant dynamical systems. At every time step $t$, the latent state of the dynamical system, ${s}_t\in\mathbb{R}^m$, stochastically generates an observation, ${x}_t\in\mathbb{R}^n$, from an observation model $P({x}_t|{s}_{t})$. The stochastic transition model $P(s_{t+1}|s_t)$ computes the predictive distribution of states at  $t+1$ given the state at time $t$. We define the belief of a latent state $s_{t+1}$ as the distribution of $s_{t+1}$ given all the past observations up to time step $t$: $\{x_1,...,x_{t}\}$, which we denote as $h_t$. 
\subsection{Belief Propagation in Latent State Space Models}
Let us define $b_t$ as the belief $P(s_t|h_{t-1})$. When the transition model $P(s_{t+1}|s_t)$ and observation model $P(x_t|s_t)$ are known, the belief $b_t$ can be computed by a special-case of message passing called forward belief propagation: 
\begin{align}
\label{eq:message_pass_HMM}
b_{t+1} =  \frac{1}{P(x_{t}|h_{t-1})}\int_{s_{t}}{b_t P(s_{t+1}|s_t)P(x_{t}|s_{t})}d s_t. 
\end{align}
%
The above equation essentially maps the belief $b_t$ and the current observation $x_t$ to the next belief $b_{t+1}$.

Consider the following linear dynamical system:
\begin{align}
\label{eq:linear_sys}
s_{t+1} = A s_{t} + \epsilon_s, \;\; \epsilon_s \sim \mathcal{N}(0, Q),\nonumber \\
x_{t} = Cs_{t} + \epsilon_x, \;\; \epsilon_x \sim \mathcal{N}(0, R),
\end{align} where $A\in\mathbb{R}^{m\times m}$ is the transition matrix, $C\in\mathbb{R}^{n\times m}$ is the observation matrix, and $Q\in\mathbb{R}^{m\times m}$ and $R\in\mathbb{R}^{n\times n}$ are noise covariances. The Kalman Filter \citep{van2012subspace} update implements the belief update in Eq.~\ref{eq:message_pass_HMM}. Since $P(s_t|h_{t-1})$ is a Gaussian distribution, we simply use the mean $\hat{s}_t$ and the covariance $\Sigma_t$ to represent $P(s_t|h_{t-1})$. The Kalman Filter update step can then be viewed as a function that maps $(\hat{s}_t, \Sigma_t)$ and the observation $x_{t}$ to $(\hat{s}_{t+1}, \Sigma_{t+1})$, which is a nonlinear map.

Given the sequences of observations $\{x_t\}_{t}$ generated from the linear dynamical system in Eq.~\ref{eq:linear_sys}, there are two common approaches to recover the parameters $A, C, Q, R$. Expectation-Maximization (EM) attempts to maximize the likelihood of the observations with respect to parameters \citep{roweis1999unifying}, but suffers from locally optimal solutions. 
The second approach relies on Spectral Learning algorithms to recover $A,C,Q,R$ up to a linear transformation \citep{van2012subspace}.\footnote{Sometimes called \emph{subspace identification}~\citep{van2012subspace} in the linear time-invariant system context.} Spectral algorithms have two key characteristics: 1) they use an observable state representation; and 2) they rely on method-of-moments for parameter identification instead of likelihood. 
Though spectral algorithms can promise global optimality in certain cases, 
this desirable property does not hold under model mismatch \citep{kulesza2014low}. In this case, using the learned parameters for filtering may result in poor filtering performance. 




\subsection{Predictive State Representations}
Recently, predictive state representations and observable operator models have been used to learn from, filter on, predict, and simulate time series data~\citep{jaeger2000observable,Littman01,Singh2004_UAI,Boots2011_IJRR,Boots11a,NIPS2015_5673}. These models provide a compact and complete description of a dynamical system that is easier to learn than latent variable models, by representing state as a set of predictions of observable quantities such as future observations. 

In this work, we follow a predictive state representation (PSR) framework and define state as the distribution of  $f_t = [x_t^T,...,x_{t+k-1}^T]^T \in\mathbb{R}^{kn}$, a $k$-step fixed-sized time window of \emph{future} observations $\{x_{t}, ..., x_{t+k-1}\}$~\citep{NIPS2015_5673}.  PSRs assume that if we can predict everything about $f_t$ at time-step $t$ (e.g., the distribution of $f_t$), then we also know everything there is to know about the state of a dynamical system at time step $t$~\citep{Singh2004_UAI}. We assume that systems we consider are $k$-observable\footnote{This assumption allows us to avoid the cryptographic hardness of the general problem~\citep{Hsu2009_COLT}.} for $k\in\mathbb{N}^+$: there is a bijective function that maps $P(s_t|h_{t-1})$ to $P(f_t|h_{t-1})$. For convenience of notation, we will present our results in terms of $k$-observable systems, where it suffices to select features from the next $k$ observations.

\begin{figure}[t]
  \centering
  \includegraphics[width=3.0in, trim={140pt 165 50 155},clip]{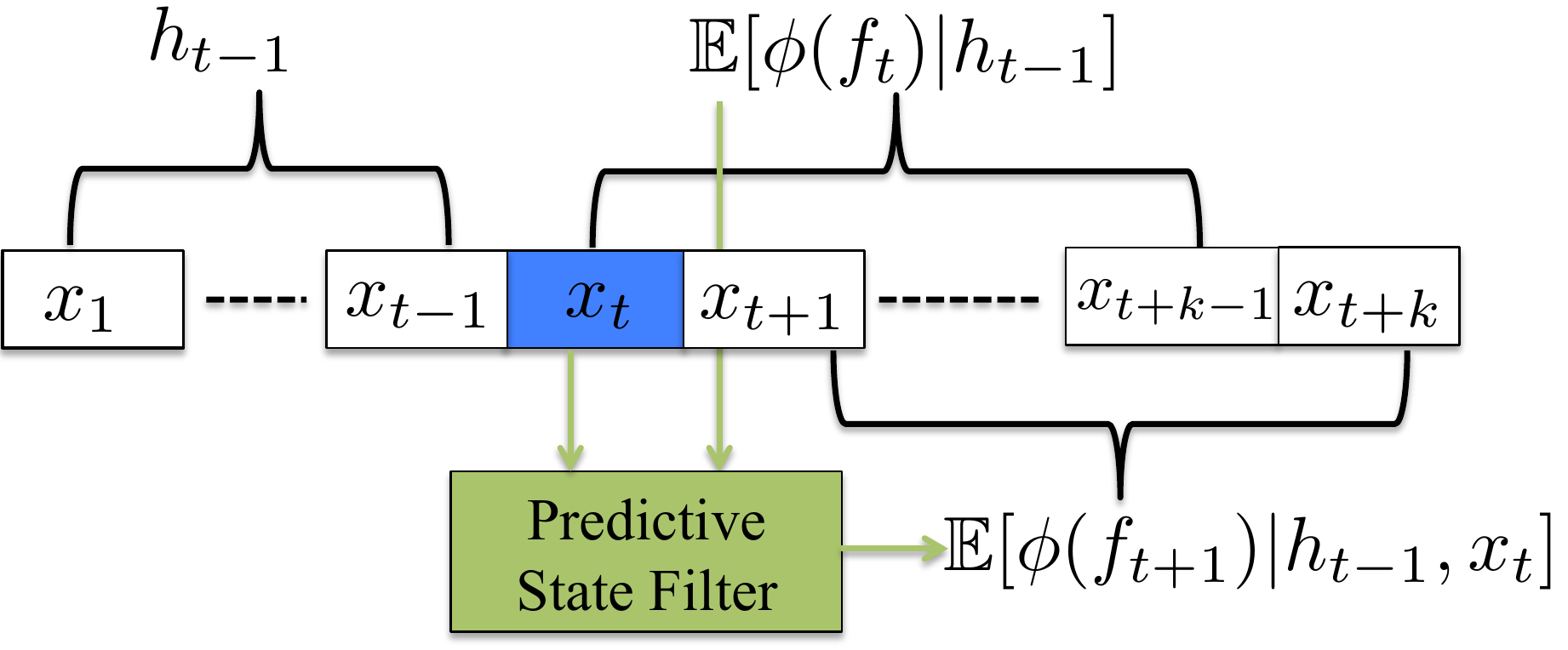}
  \vspace{-2mm}
  \caption{Filtering with predictive states for a $k$-observable system. At time step $t$, the filter uses the belief $\mathbb{E}[\phi(f_t)|h_{t-1}]$ and the latest observation $x_t$ as feedback, outputs the next belief $\mathbb{E}[\phi(f_{t+1})|h_{t-1},x_t]$. }
\vspace{-3mm}
\label{fig:ps_filter}
\end{figure}

Following~\citet{NIPS2015_5673}, we define the predictive state at time step $t$ as $\mathbb{E}[\phi(f_t) | h_{t-1}]$ where $\phi$ is some feature function that is sufficient for the distribution $P(f_t|h_{t-1})$. The expectation is taken with respect to the distribution $P(f_t|h_{t-1})$: $\mathbb{E}[\phi(f_t) | h_{t-1}] = \int_{f_t} \phi(f_t)P(f_t|h_{t-1})d f_t $. The conditional expectation can be understood as a function of which the input is the random variable $h_{t-1}$. For example, we can set $\mathbb{E}[\phi(f)|h_{t-1}] = \mathbb{E}[f, ff^T|h_{t-1}]$ if $P(f_t|h_{t-1})$ is a Gaussian distribution (e.g., linear dynamical system in Eq.~\ref{eq:linear_sys} ); or we can set $\phi(f) = [x_{t}\otimes...\otimes x_{t+k-1}]$ if we are working on a discrete models (discrete latent states and discrete observations), where $x_t$ is an indicator vector  representation of the observation and $\otimes$ is the tensor product. Therefore, we assume that there exists a bijective function mapping $P(f|h_{t-1})$ to $\mathbb{E}[\phi(f_t)|h_{t-1}]$. For any test $f'_t$, we can compute the probability of $P(f'_t|h_{t-1})$ by simply using the predictive state $\mathbb{E}[\phi(f_t)|h_{t-1}]$. Note that the mapping from $\mathbb{E}[\phi(f_t)|h_{t-1}]$ to $P(f_t'|h_{t-1})$ is not necessarily linear. 

To filter from the current predictive state $\mathbb{E}[\phi(f_t)|h_{t-1}]$ to the next state $\mathbb{E}[\phi(f_{t+1})|h_{t}]$ conditioned on the most recent observation $x_t$ (see Fig.~\ref{fig:ps_filter} for an illustration),
PSRs additionally define an extended state 
$\mathbb{E}[\zeta(f_t, x_{t+k}) | h_{t-1}] =  \int_{(f_t,x_{t+k})}\zeta(f_t,x_{t+k})P(f_t,x_{t+k}|h_{t-1})d(f_t,x_{t+k})$, where $\zeta$ is another feature function for the future observations $f_t$ and one more observation $x_{t+k}$. PSRs explicitly assume there exists a linear relationship between $\mathbb{E}[\phi(f_t)|h_{t-1}]$ and $\mathbb{E}[\zeta(f_t,x_{t+k})|h_{t-1}]$, which can be learned by \emph{Instrumental Variable Regression} (IVR) \citep{NIPS2015_5673}. PSRs then additionally assume a nonlinear conditioning operator 
that can compute the next predictive state with the extended state and the latest observation as inputs. 


\section{Predictive State Inference Machines}
\label{sec:psim}
The original Inference Machine framework 
reduces the problem of learning graphical models to solving a set of classification or regression problems, where the learned classifiers mimic message passing procedures that output marginal distributions for the nodes in the model 
\citep{langford2009learning,ross2011_CVPR, bagnelllearning}. However, Inference Machines cannot be applied to learning latent state space models (unsupervised learning) since we do not have access to hidden states' information. 


We tackle this problem with predictive states. By using an observable representation for state, observations in the training data can be used for supervision in the inference machine. More formally, instead of tracking the hidden state $s_t$, we focus on the corresponding predictive state $\mathbb{E}[\phi(f_t)|h_{t-1}]$. Assuming that the given predictive state $\mathbb{E}[\phi(f_t)|h_{t-1}]$ can reveal the probability $P(f_t | h_{t-1})$, we use the training data $f_t$ 
to quantify how good the predictive state is by computing the likelihood of $f_t$. 
The goal is to learn an operator ${F}$ (the green box in Fig.~\ref{fig:ps_filter}) which \emph{deterministically} passes the predictive states forward in time conditioned on the latest observation:
\begin{align}
\mathbb{E}[\phi(f_{t+1})|h_{t}] = {F}\Big(\mathbb{E}[\phi(f_t)|h_{t-1}], x_t\Big),
\end{align} such that the likelihood of the observations $\{f_t\}_{t}$ being generated from the sequence of predictive states $\{\mathbb{E}[\phi(f_t)|h_{t-1}]\}_{t}$ is maximized. In the standard PSR framework, the predictor $F$ can be regarded as the composition of the linear mapping (from predictive state to extended state) and the conditioning operator. Below we show if we can correctly filter with predictive states, then this is equivalent to filtering with latent states as in Eq.~\ref{eq:message_pass_HMM}.

\subsection{Predictive State Propagation}
The belief propagation in Eq.~\ref{eq:message_pass_HMM} is for latent states $s_t$. We now describe the corresponding belief propagation for  updating the predictive state from $\mathbb{E}[\phi(f_t)|h_{t-1}]$ to $\mathbb{E}[\phi(f_{t+1})|h_t]$ conditioned on the new observation $x_t$. Since we assume that the mapping from $P(s_t|h_{t-1})$ to $P(f_t|h_{t-1})$ and the mapping from $P(f_t|h_{t-1})$ to $\mathbb{E}[\phi(f_t)|h_{t-1}]$ are both bijective, there must exist a bijective map $\mathrm{q}$ and its inverse $\mathrm{q}^{-1}$ such that $\mathrm{q}(P(s_t|h_{t-1})) = \mathbb{E}[\phi(f_t)|h_{t-1}]$ and $\mathrm{q}^{-1}(\mathbb{E}[\phi(f_t)|h_{t-1}]) = P(s_t|h_{t-1})$,\footnote{The composition of two bijective functions is bijective.} then the message passing in Eq.~\ref{eq:message_pass_HMM} is also equivalent to:
\begin{align}
\label{eq:message_pass_HMM_PSR}
&\mathbb{E}[\phi(f_{t+1})|h_{t}] = \mathrm{q}(P(s_{t+1}|{h_{t}}))  \\
& = \mathrm{q}\big( \int_{s_t}\frac{P(s_{t}|h_{t-1})P(s_{t+1}|s_t)P(x_{t}|s_{t})}{P(x_{t}|h_{t-1})}d s_t   \big) \nonumber \\
& = \mathrm{q}\big(\int_{s_t}\frac{\mathrm{q}^{-1}(\mathbb{E}[\phi(f_t)|h_{t-1}]) P(s_{t+1}|s_t)P(x_{t}|s_{t}) }{P(x_{t}|h_{t-1})}d{s_t}\big) \nonumber 
\end{align}%
Eq.~\ref{eq:message_pass_HMM_PSR} explicitly defines the map ${F}$ that takes the inputs of $\mathbb{E}[\phi(f_t)|h_{t-1}]$ and $x_t$ and outputs $\mathbb{E}[\phi(f_{t+1})|h_t]$. This map $F$ could be non-linear since it depends the transition model $P(s_{t+1}|s_t)$, observation model $P(x_t|s_t)$ and function $\mathrm{q}$, which are all often complicated, non-linear functions in real dynamical systems. We do not place any parametrization assumptions on the transition and observation models. Instead, we parametrize and restrict the class of predictors to encode the underlying dynamical system and aim to find a predictor $F$ from the restricted class. We call this framework for inference the \mbox{\PBIM (\pbim).}

\pbim is different from PSRs in the following respects: (1) \pbim collapses the two steps of PSRs (predict the extended state and then condition on the latest observation) into one step---as an Inference Machine---for closed-loop update of predictive states; (2) \pbim directly targets the filtering task and has theoretical guarantees on the filtering performance; (3) unlike PSRs where one usually needs to utilize linear PSRs for learning purposes \citep{Boots2011_IJRR},  \pbim can generalize to non-linear dynamics by leveraging non-linear regression or classification models.  

Imagine that we can perform belief propagation with \pbim in predictive state space as shown in Eq.~\ref{eq:message_pass_HMM_PSR}, then this is \emph{equivalent} to classic filter with latent states as shown in Eq.~\ref{eq:message_pass_HMM}. To see this, we can simply apply $\mathrm{q}^{-1}$ on both sides of the above equation Eq.~\ref{eq:message_pass_HMM_PSR}, which exactly reveals Eq.~\ref{eq:message_pass_HMM}. 
We refer readers to the Appendix for a detailed case study of the stationary Kalman Filter, where we explicitly show this equivalence. Thanks to this equivalence, we can learn accurate inference machines, even for partially observable systems. We now turn our focus on learning the map ${F}$ in the predictive state space. 


\subsection{Learning Non-stationary Filters with Predictive States}
For notational simplicity, let us define trajectory as $\tau$, which is sampled from a unknown distribution $\mathcal{D}_{\tau}$. We denote the predictive state as $m_t = \mathbb{E}[\phi(f_t)|h_{t-1}]$. We use $\hat{m}_t$ to denote an approximation of $m_t$. Given a predictive state $m_t$ and a noisy observation $f_t$ conditioned on the history $h_{t-1}$, we let the loss function\footnote{Squared loss in an example Bregman divergence of which there are others that are optimized by the conditional expectation~\citep{banerjee2005}. We can design $d(m_t,f_t)$ as negative log-likelihood, as long as it can be represented as a Bregman divergence (e.g., negative log-likelihood of distributions in exponential family).} $d(m_t, f_t) = \|m_t - \phi(f_t)\|_2^2$. This squares loss function can be regarded as matching moments. For instance, in the stationary Kalman filter setting, we could set $m_t = \mathbb{E}[f_t|h_{t-1}]$ and $d(m_t, f_t) = \|m_t - f_t\|_2^2$ (matching the first moment). 


We first present a algorithm for learning non-stationary filters using  \emph{Forward Training} \citep{ross2010efficient} in Alg.~\ref{alg:PBIM_forward_training}. Forward Training learns a non-stationary filter for each time step. Namely, at time step $t$, forward training learns a hypothesis $F_t$ that approximates the filtering procedure at time step $t$:
$\hat{m}_{t+1} = F_t(\hat{m}_t, x_t)$,
 where $\hat{m}_t$ is computed by $F_{t-1}(\hat{m}_{t-1}, x_{t-1})$ and so on. 
Let us define $\hat{m}_t^i$ as the predictive state computed by rolling out $F_1,..,F_{t-1}$ on trajectory $\tau_i$ to time step $t-1$. We define $f_{t}^i$ as the next $k$ observations starting at time step $t$ on trajectory $\tau_i$. At each time step $t$, the algorithm collects a set of training data $D_t$, where the feature variables $z_t$ consist of the predictive states $\hat{m}_{t}^i$ from the previous hypothesis $F_{t-1}$ and the local observations $x_t^i$, and the targets consist of the corresponding future observations $f_{t+1}^i$ across all trajectories $\tau_i$. It then trains a new hypothesis $F_{t}$ over the hypothesis class $\mathcal{F}$ to minimize the loss over dataset $D_t$.

\begin{algorithm}[tb]
\caption{\PBIM (\pbim) with Forward Training}
 \label{alg:PBIM_forward_training}
\begin{algorithmic}[1]
  \STATE {\bfseries Input:} $M$ independent trajectories $\tau_i$, $1\leq i\leq M$;
  \STATE Set $\hat{m}_1 = \frac{1}{M}\sum_{i=1}^M \phi(f^i_1)$; 
 \STATE Set $\hat{m}^i_1 = \hat{m}_1$ for trajectory $\tau_i, 1\leq i\leq M$;
 \FOR {t = 1 to T}
    \STATE For each trajectory $\tau_i$, add the input $z_t^i = (\hat{m}^i_t, x^i_t)$ to $D_t$ as feature variables and the corresponding $f^i_{t+1}$ to $D_t$ as the targets;
    \STATE Train a hypothesis $F_t$ on $D_t$ to minimize the loss $d(F(z), f)$ over $D_t$; 
    \STATE For each trajectory $\tau_i$, roll out $F_1,...,F_t$ along the trajectory (Eq.~\ref{eq:roll_out_2}) to compute $\hat{m}_{t+1}^{i}$;
 \ENDFOR
\STATE {\bfseries Return:} the sequence of hypothesis $\{F_t\}_{t=1}^N$.
\end{algorithmic}
\end{algorithm}

\pbim with Forward Training aims to find a good sequence of hypotheses $\{F_t\}$ such that:
\begin{align}
\label{eq:filter_error_ft}
&\min_{F_1\in\mathcal{F},...F_T\in\mathcal{F}} \mathbb{E}_{\tau\sim \mathcal{D}_{\tau}} \Big[ \frac{1}{T}\sum_{t=1}^T d(F_t(\hat{m}_t^{\tau}, x_t^{\tau}), f_{t+1}^{\tau}) \Big], \\
&  \;\;\;\;\;\;\; s.t. \;\; \hat{m}_{t+1}^{\tau} = F_t(\hat{m}_t^{\tau}, x_t^{\tau}), \forall t\in [1,T-1],
\label{eq:roll_out_2}
\end{align} where $\hat{m}_1 = \arg\min_m \sum_{t=1}^M d(m, f_1^i)$, which is equal to $\frac{1}{T}\sum_{i=1}^T \phi(f_i^i)$.
Let us define $\omega_{t}$ as the joint distribution of feature variables $z_{t}$ and targets $f_{t+1}$ after rolling out $F_1,..., F_{t-1}$ on the trajectories sampled from $\mathcal{D}_{\tau}$. Under this definition, the filter error defined above is equivalent to $\frac{1}{T}\sum_{t=1}^T\mathbb{E}_{(z,f)\sim \omega_{t}}  \Big[ d(F_t(z), f) \Big]$. 
Note essentially the dataset $D_t$ collected by Alg.~\ref{alg:PBIM_forward_training} at time step $t$ forms a finite sample estimation of $\omega_{t}$.  

To analyze the consistency of our algorithm, we assume every learning problem can be solved perfectly (risk minimizer finds the Bayes optimal) \cite{langford2009learning}. We first show that under infinite many training trajectories, and in realizable case --- the underlying true filters $F_1^*, ..., F_T^*$ are in the hypothesis class $\mathcal{F}$,  Alg.~\ref{alg:PBIM_forward_training} is consistent:
\begin{theorem}
\label{them:consistent}
With infinite many training trajectories and in the realizable case, if all learning problems are solved perfectly,  the sequence of predictors $F_1, F_2, ..., F_T$ from Alg.~\ref{alg:PBIM_forward_training} can generate exact predictive states $\mathbb{E}[\phi(f_t^{\tau})|h_{t-1}^{\tau}]$ for any trajectory $\tau\sim\mathcal{D}_{\tau}$ and $1\leq t\leq T$.
\end{theorem} We include all proofs in the appendix. Next for the agnostic case, we show that Alg.~\ref{alg:PBIM_forward_training} can still achieve a reasonable upper bound. Let us define $\epsilon_t  = \min_{F\sim\mathcal{F}}\mathbb{E}_{(z,f)\sim \omega_t}[d(F(z), f)]$, which is the minimum batch training error under the distribution of inputs resulting from hypothesis class $\mathcal{F}$. Let us define $\epsilon_{\max} = \max_{t}\{\epsilon_t\}$. Under infinite many training trajectories, even in the model agnostic case, we have the following guarantees for filtering error for Alg.~\ref{alg:PBIM_forward_training}:
\begin{theorem}
\label{them:forward_train_infinite}
With infinite many training trajectories, for the sequence $\{F_t\}_t$ generated by Alg.~\ref{alg:PBIM_forward_training}, we have:
\begin{equation}
\mathbb{E}_{\tau\sim\mathcal{D}_{\tau}}\Big[  \frac{1}{T}\sum_{t=1}^Td(F_t(\hat{m}_t^{\tau}, x_t^{\tau}), f_{t+1}^{\tau})  \Big] = \frac{1}{T}\sum_{t}\epsilon_t \leq \epsilon_{max}. \nonumber
\end{equation} 
\end{theorem}
Theorem.~\ref{them:forward_train_infinite} shows that the filtering error is upper-bounded by the average of the minimum batch training errors from each step. If we have a rich class of hypotheses and small noise (e.g., small Bayes error), $\epsilon_t$ could be small. 

To analyze finite sample complexity, we need to split the dataset into $T$ disjoint sets to make sure that the samples in the dataset $D_t$ are i.i.d (see details in Appendix). Hence we reduce forward training to $T$ independent passive supervised learning problems. We have the following agnostic theoretical bound:
\begin{theorem}
\label{them:finite_sample_analysis_forward_training}
With $M$ training trajectories, for any $F_t^*\in\mathcal{F}, \forall t$, we have with probability at least $1-\delta$:
\begin{small}
\begin{align}
&\mathbb{E}_{\tau\sim\mathcal{D}_{\tau}}\Big[  \frac{1}{T}\sum_{t=1}^Td(F_t(\hat{m}_t^{\tau}, x_t^{\tau}), f_{t+1}^{\tau})  \Big]  \nonumber \\
& \leq \mathbb{E}_{\tau\sim \mathcal{D}_{\tau}}\big[ \frac{1}{T}\sum_{t=1}^T d(F_t^*(\hat{m}_t^{\tau}, x_t^{\tau}), f_{t+1}^{\tau})   \big] \nonumber \\
&\;\;\;\;\;\;\;\;+4\nu\bar{\mathcal{R}}(\mathcal{F})  + 2\sqrt{\frac{T\ln(T/\delta)}{2M}},
\end{align}\end{small}%
where \small{$v = \sup_{F,z,f} 2\|F(z) - f\|_2$,  $\bar{\mathcal{R}}(\mathcal{F}) = \frac{1}{T}\sum_{t=1}^T\mathcal{R}_t(\mathcal{F}))$} and $\mathcal{R}_t(\mathcal{F})$ is the Rademacher number of $\mathcal{F}$ under $\omega_t$. 
\label{them:forward_train_finite}
\end{theorem}
As one might expect, the learning problem becomes harder as $T$ increases, however our finite sample analysis shows the average filtering error grows sublinearly as $\tilde{O}(\sqrt{T})$. 

Although Alg.~\ref{alg:PBIM_forward_training} has nice theoretical properties, one shortcoming is that it is not very data efficient. In practice, it is possible that we only have small number of training trajectories but each trajectory is long ($T$ is big). This means that we may have few training data samples (equal to the number of trajectories) for learning hypothesis $F_t$. Also, instead of learning non-stationary filters, we often prefer to learn a stationary filter such that we can filter indefinitely. In the next section, we present a different algorithm that utilizes \emph{all} of the training data to learn a stationary filter.


\subsection{Learning Stationary Filters with Predictive States}
\label{sec:DAgger}
The optimization framework for finding a good stationary filter $F$ is defined as:
\begin{align}
\label{eq:objective}
&\min_{{F}\in\mathcal{F}}\mathbb{E}_{\tau\sim \mathcal{D}_{\tau}}\frac{1}{T}\sum_{t=1}^{T} d(F(\hat{m}_{t}, x_{t}), f_{t+1}), \\
&s.t \;\;\;\; \hat{m}_{t+1} = F(\hat{m}_{t}, x_t), \forall t\in[1,T-1],
\label{eq:roll_out}
\end{align} where $\hat{m}_1 = \arg\min_m \sum_{t=1}^M d(m, f_1^i) = \frac{1}{T}\sum_{i=1}^T \phi(f_i^i)$.
Note that the above objective function is non-convex, since $\hat{m}_t$ is computed recursively and in fact is equal to $F(...F(F(\hat{m}_1, x_1),x_2)...)$, where we have $t-1$ nested $F$. As we show experimentally, optimizing this objective function via Back-Propagation likely leads to local optima. Instead, we optimize the above objective function using an iterative approach called Dataset Aggregation (DAgger) \cite{Ross2011_AISTATS} (Alg.~\ref{alg:hidden_inference_machine}). 
Due to the non-convexity of the objective, DAgger also will not promise global optimality. But as we will show, \pbim with DAgger gives us a sound theoretical bound for filtering error. 

Given a trajectory $\tau$ and hypothesis $F$, we define $\hat{m}^{\tau, F}_t$ as the predictive belief generated by $F$ on $\tau$ at time step $t$. We also define $z^{\tau, F}_{t}$ to represent the feature variables $(\hat{m}^{\tau,F}_t, x^{\tau}_t)$. 
At iteration $n$, Alg.~\ref{alg:hidden_inference_machine} rolls out the predictive states using its current hypothesis $F_{n}$ (Eq.~\ref{eq:roll_out}) on all the given training trajectories (Line.~\ref{line:rolling_out}). Then it collects all the feature variables $\{(\hat{m}^{i,F_n}_t,x^i_t)\}_{t,i}$ and the corresponding target variables $\{f^i_{t+1}\}_{t,i}$ to form a new dataset $D_n'$ and aggregates it to the original dataset $D_{n-1}$. Then a new hypothesis $F_{n}$ is learned from the aggregated dataset $D_n$ by minimizing the loss $d(F(z),f)$ over $D_n$. 

\begin{algorithm}[tb]
 \caption{\PBIM (\pbim) with DAgger}
 \label{alg:hidden_inference_machine}
 \begin{algorithmic}[1]
 \STATE {\bfseries Input:} $M$ independent trajectories $\tau_i$, $1\leq i\leq M$;
 \STATE Initialize $D_0 \leftarrow\emptyset $ and initalize $F_0$ to be any hypothesis in $\mathcal{F}$;
 \STATE Initialize $\hat{m}_1 = \frac{1}{M}\sum_{i=1}^M \phi(f^i_1)$\;
 \FOR{n = 0 to N}
    \STATE \label{line:rolling_out} Use $F_{n}$ to perform belief propagation (Eq.~\ref{eq:roll_out}) on trajectory $\tau_i$, $1\leq i\leq M $
    \STATE For each trajectory $\tau_i$ and each time step $t$, add the input $z^i_t = ({m}^{i,F_n}_t, x^i_{t})$ encountered by $F_{n}$ to $D_{n+1}'$ as feature variables and the corresponding $f^i_{t+1}$ to $D_{n+1}'$ as the targets  \label{line:collect};
    \STATE Aggregate dataset $D_{n+1} = D_{n}\cup D_{n+1}'$;
    \STATE Train a new hypothesis $F_{n+1}$ on $D_{n+1}$ to minimize the loss $d(F(m,x), f)$\label{line:batch_learn}; 
 \ENDFOR
 \STATE {\bfseries Return:} the best hypothesis $\hat{F}\in\{F_n\}_n$ on validation trajectories.
 \end{algorithmic}
\end{algorithm}

\begin{table*}[t]
\begin{center}
\resizebox{0.9\textwidth}{!}{ 
    \begin{tabular}{| l | l | l | l | l | l | l| }
    \hline
      & \textbf{N4SID} & \textbf{IVR} & \textbf{\pbim-Linear}$_d$ &\textbf{\pbim-Linear$_b$}  & \textbf{\pbim-RFF}$_d$ & \textbf{Traj. Pwr} \\ \hline
    \textbf{Robot Drill Assembly} & 2.87$\pm$0.2 & 2.39 $\pm$0.1 & 2.15$\pm$0.1 &  2.54$\pm$0.1 & \textbf{1.80$\pm$0.1}  &27.90\\ \hline 
    \textbf{Motion Capture} & 7.86$\pm$ 0.8 & 6.88$\pm$ 0.7 & 5.75$\pm$0.5 & 9.94$\pm$2.9 &\textbf{5.41$\pm$ 0.5} &  107.92 \\ \hline
    \textbf{Beach Video Texture} & 231.33$\pm$10.5 & 213.27$\pm$11.5 & 164.23$\pm$8.7 & 268.73$\pm$9.5  & \textbf{130.53$\pm$9.1} & 873.77 \\ \hline
    \textbf{Flag Video Texture} & 3.38e3$\pm$1.2e2 & 3.38e3$\pm$1.3e2 & 1.28e3$\pm$7.1e1 & 1.31e3$\pm$6.4e1  & \textbf{1.24e3$\pm$9.6e1} & 3.73e3 \\ \hline
    \end{tabular}}
\end{center}
\vspace{-4mm}
\caption{Filter error (1-step ahead) and standard deviation on different datasets. We see that using \pbim with DAgger with both RFF and Linear outperforms the spectral methods N4SID and IVR, with the RFF performing better on almost all the datasets. DAgger (20 iterations) trains a better linear regression for \pbim than back-propagation with random initialization (400 epochs). We also give the average trajectory power for the true observations from each dataset. \vspace{-2mm}
}
\label{tab:mean_std}
\end{table*}

Alg.~\ref{alg:hidden_inference_machine} essentially utilizes DAgger to optimize the non-convex objective in Eq.~\ref{eq:objective}. By using DAgger, we can guarantee a hypothesis that, when used during filtering, performs nearly as well as when performing regression on the aggregate dataset $D_N$. In practice, with a rich hypothesis class $\mathcal{F}$ and small noise (e.g., small Bayes error), small regression error is possible. We now analyze the filtering performance of \pbim with DAgger below.

Let us fix a hypothesis $F$ and a trajectory $\tau$, we define $\omega_{F,\tau}$ as the uniform distribution of $(z, f)$: $\omega_{F,\tau} = \mathcal{U}\Big[ (z_1^{\tau,F}, f^{\tau}_2), ..., (z_T^{\tau,F}, f^{\tau}_{T+1})   \Big]$. Now we can rewrite the filtering error in Eq.~\ref{eq:objective} as $L(F) = \mathbb{E}_{\tau}[\mathbb{E}_{z,f\sim \omega_{F,\tau}}[d(F(z), f) ]|\tau]$.
Let us define the loss function for any predictor $F$ at iteration $n$ of Alg.~\ref{alg:hidden_inference_machine} as:
\begin{align}
L_n(F) = \mathbb{E}_{\tau} [\mathbb{E}_{z,f\sim \omega_{F_n,\tau}} [d(F(z), f)]|\tau].
\end{align} As we can see, at iteration $n$, the dataset $D'_n$ that we collect forms an empirical estimate of the loss $L_n$:
\begin{align}
\hat{L}_n(F)
& = \frac{1}{M}\sum_{\tau = 1}^M \mathbb{E}_{z,f\sim \omega_{F_n, \tau}}\big(d(F(z), f)\big).
\end{align} 
We first analyze the algorithm under the assumption that $M=\infty,$ $\hat{L}_n(F) = L_n(F)$. Let us define \emph{Regret} $\gamma_N$ as:
$\frac{1}{N}\sum_{n=1}^N L_n(F_n) - \min_{F\in\mathcal{F}}\frac{1}{N}\sum_{n=1}^N L_n(F) \leq \gamma_N$.
 We also define the minimum average training error $\epsilon_N = \min_{F\in\mathcal{F}}\frac{1}{N} \sum_{n=1}^N L_n(F)$. Alg.~\ref{alg:hidden_inference_machine} can be regarded as running the Follow the Leader (FTL) \cite{cesa2004,shalev2009mind,hazan2007logarithmic} on the sequence of loss functions $\{L_n(F)\}_{n=1}^N$.  When the loss function $L_n(F)$ is strongly convex with respect to $F$, FTL is no-regret in a sense that $\lim_{N\to\infty}\gamma_N = 0$. 
Applying Theorem 4.1 and its reduction to no-regret learning analysis from \cite{Ross2011_AISTATS} to our setting, we have the following guarantee for filtering error:
\begin{corollary}
\label{them:infinite}
\citep{Ross2011_AISTATS} For Alg.~\ref{alg:hidden_inference_machine},  there exists a predictor $\hat{F}\in\{F_n\}_{n=1}^N$ such that:
\begin{align}
L(\hat{F}) = \mathbb{E}_{\tau}\big[\mathbb{E}_{z,f\sim \omega_{\hat{F},\tau}}(d(\hat{F}(z), f))|\tau \big] \leq \gamma_N + \epsilon_N. \nonumber
\end{align}
\end{corollary}
As we can see, under the assumption that $L_n$ is strongly convex, as $N\to\infty$, $\gamma_N$ goes to zero. Hence the filtering error of $\hat{F}$ is upper bounded by the minimum batch training error that could be achieved by doing regression on $D_N$ within class $\mathcal{F}$. 
In general the term $\epsilon_N$ depends on the noise of the data and the expressiveness of the hypothesis class $\mathcal{F}$. Corollary.~\ref{them:infinite} also shows for fully realizable and noise-free case, \pbim with DAgger finds the optimal filter that drives the filtering error to zero when $N\rightarrow \infty$. 



The finite sample analysis from \cite{Ross2011_AISTATS} can also be applied to \pbim. Let us define $\hat{\epsilon}_N = \min_{F\in\mathcal{F}}\frac{1}{N}\hat{L}_n(F)$,  $\hat{\gamma}_N \geq \frac{1}{N}\sum_{n=1}^N \hat{L}_n(F_n) - \min_{F\in\mathcal{F}}\frac{1}{N}\sum_{n=1}^N \hat{L}_n(F)$, we have: 
\begin{corollary}
\citep{Ross2011_AISTATS} For Alg.~\ref{alg:hidden_inference_machine},  there exists a predictor $\hat{F}\in\{F_n\}_{n=1}^N$ such that with probability at least $1-\delta$:
\begin{small}
\begin{align}
&L(\hat{F}) = \mathbb{E}_{\tau}\big[\mathbb{E}_{z,f\sim \omega_{\hat{F},\tau}}(d(\hat{F}(z), f))|\tau \big] \leq \hat{\gamma}_N + \hat{\epsilon}_N \nonumber \\
&\;\;\;\;\;\; + L_{\mathrm{max}}(\sqrt{\frac{2\ln (1/\delta)}{MN}}).
\end{align} \end{small}%
\end{corollary}




\vspace{-10pt}
\section{Experiments}
\label{sec:exp}
We evaluate the \pbim on a variety of dynamical system benchmarks. We use two feature functions: $\phi_1(f_t) = [x_t,..., x_{t+k-1}]$, which stack the $k$ future observations together (hence the message $m$ can be regarded as a prediction of future $k$ observations $(\hat{x}_t,..,\hat{x}_{t+k-1})$), and $\phi_2(f_t) = [x_t,...,x_{t+k-1}, {x_t}^2,...,x_{t+k-1}^2]$, which includes second moments (hence $m$ represents a Gaussian distribution approximating the true distribution of future observations). To measure how good the computed predictive states are, we extract $\hat{x}_i$ from $\hat{m}_t$, and evaluate $\|\hat{x}_i - x_i\|_2^2$, 
 the squared distance between the predicted observation $\hat{x}_i$ and the corresponding true observation $x_i$. 
We implement \pbim with DAgger using two underlying regression methods: ridge linear regression (\textbf{\pbim-Linear}$_d$) and  linear ridge regression with Random Fourier Features (\textbf{\pbim-RFF}$_d$) \citep{rahimi2007}\footnote{With RFF, \pbim approximately embeds the distribution of $f_t$ into a Reproducing Kernel Hilbert Space.}. We also test \pbim with back-propagation for linear regression (\textbf{\pbim-Linear}$_b$). We compare our approaches to several baselines: Autoregressive models (\textbf{AR}), Subspace State Space System Identification (\textbf{N4SID})~\citep{van2012subspace}, and PSRs implemented with {\textbf{IVR}~\citep{NIPS2015_5673}.

\subsection{Synthetic Linear Dynamical System}
First we tested our algorithms on a synthetic
linear dynamical system (Eq.~\ref{eq:linear_sys}) with a 2-dimensional observation $x$. We designed the system such that it is exactly $2$-observable.  The sequences of observations are collected from the linear stationary Kalman filter of the LDS \citep{Boots2012_Thesis,NIPS2015_5673}. The details of the LDS are in Appendix. 
\begin{figure}[t]
  \centering
  \vspace{-4mm}
  \begin{overpic}[width=2.75in]{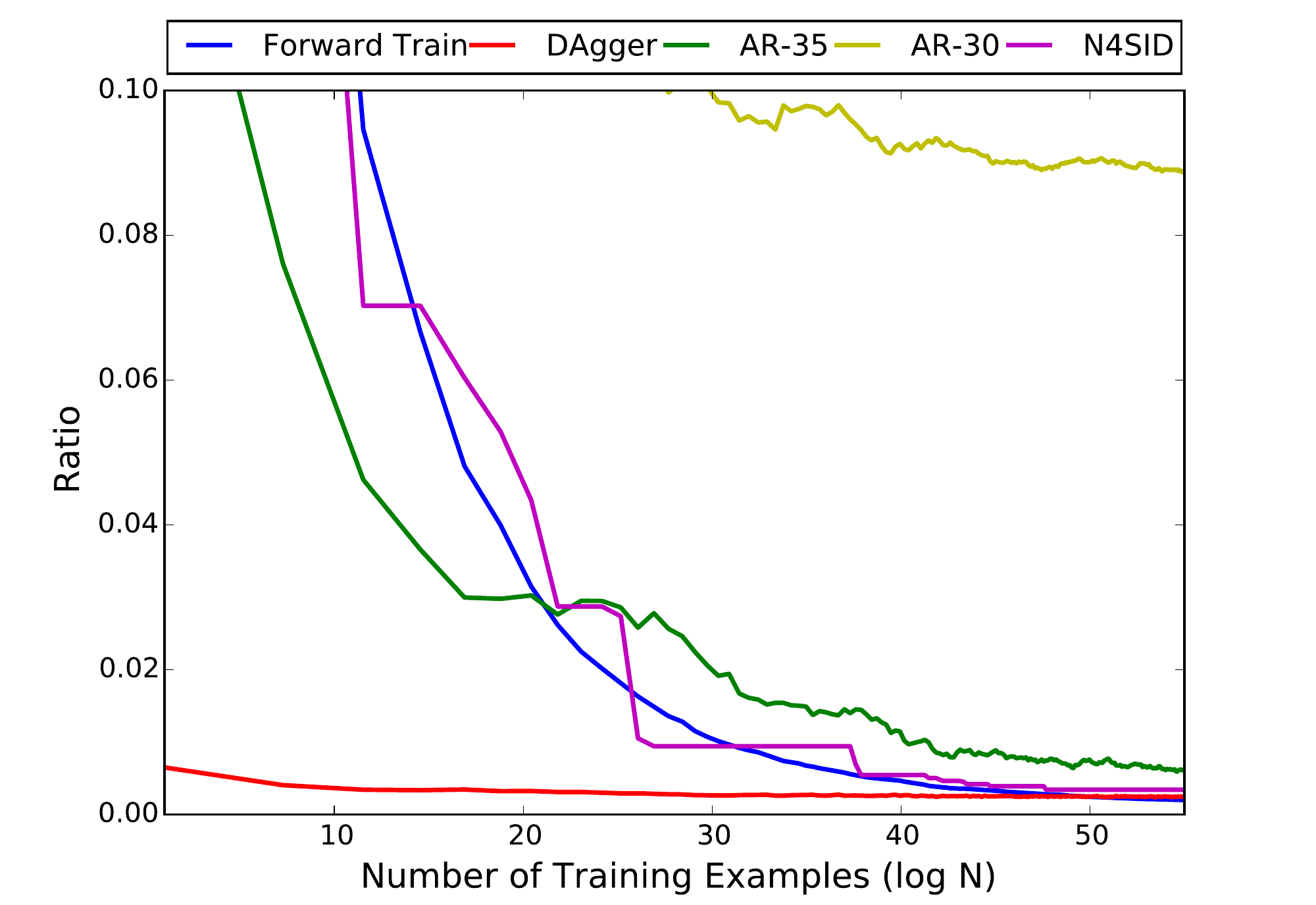}
  \end{overpic}
  \vspace{-2mm}
  \caption{The convergence rate of different algorithms. The ratios (y-axis) are computed as $\log(\frac{e}{e_F})$ for error $e$ from corresponding algorithms. The x-axis is computed as $\log(N)$, where $N$ is the number of trajectories used for training.\vspace{-4mm}}
  \label{fig:forward_training_convergence}
\end{figure}

\begin{figure*}[t!]
	\centering
	\vspace{-2mm}
	\begin{subfigure}[l]{0.3\textwidth}
        \includegraphics[width=1.0\textwidth,keepaspectratio]{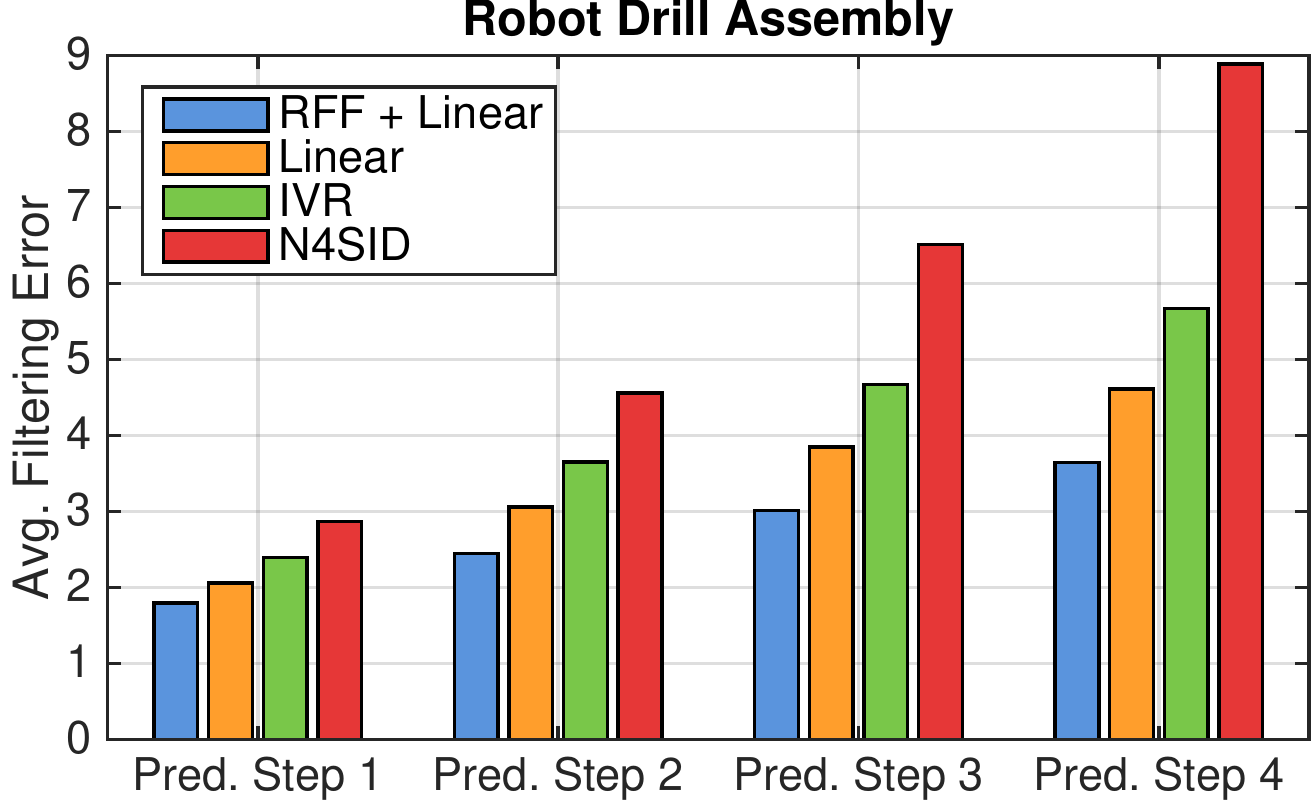}
    	\caption{Robot Drill Assembly}
    \end{subfigure}
	\begin{subfigure}[l]{0.3\textwidth}
        \includegraphics[width=1.0\textwidth,keepaspectratio]{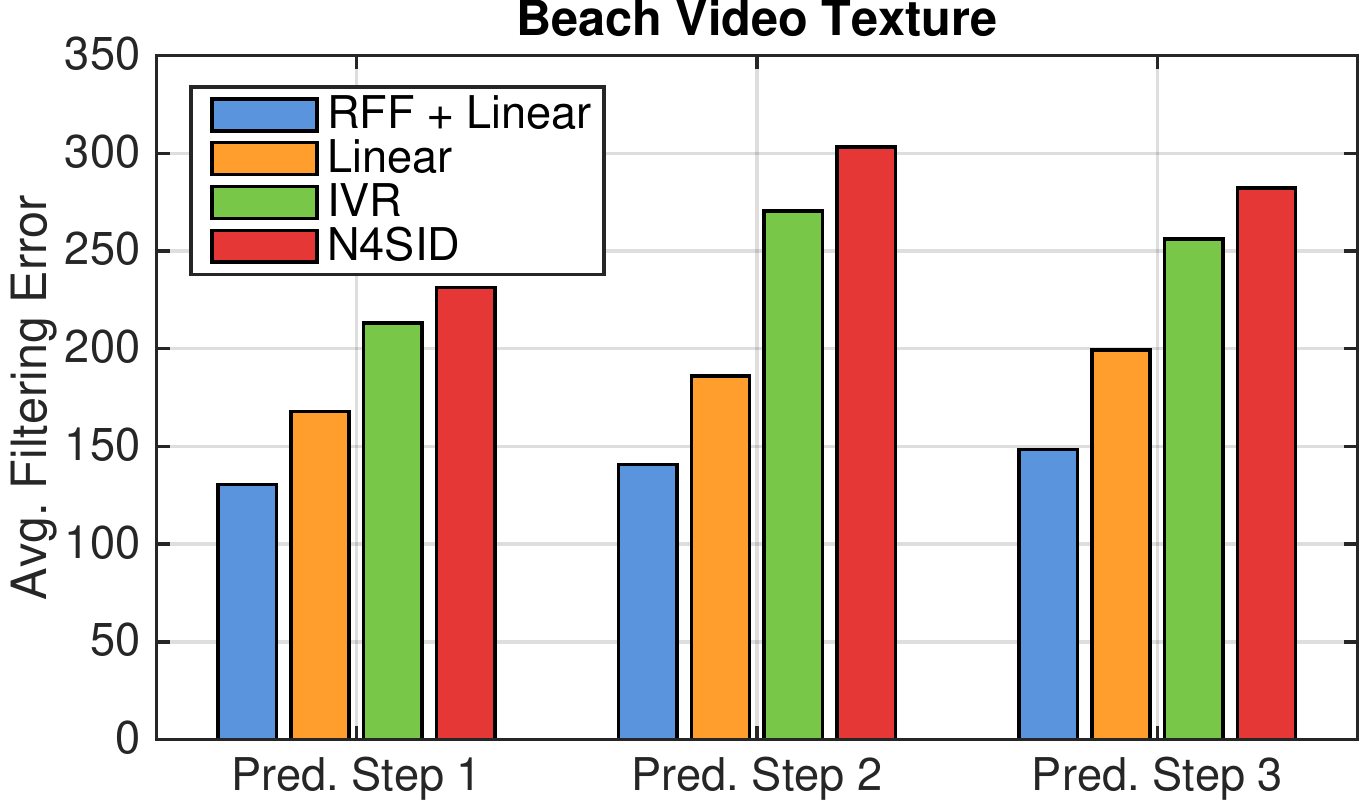}
    	\caption{Beach Video Texture}
    \end{subfigure}
	\begin{subfigure}[l]{0.3\textwidth}
        \includegraphics[width=1.0\textwidth,keepaspectratio]{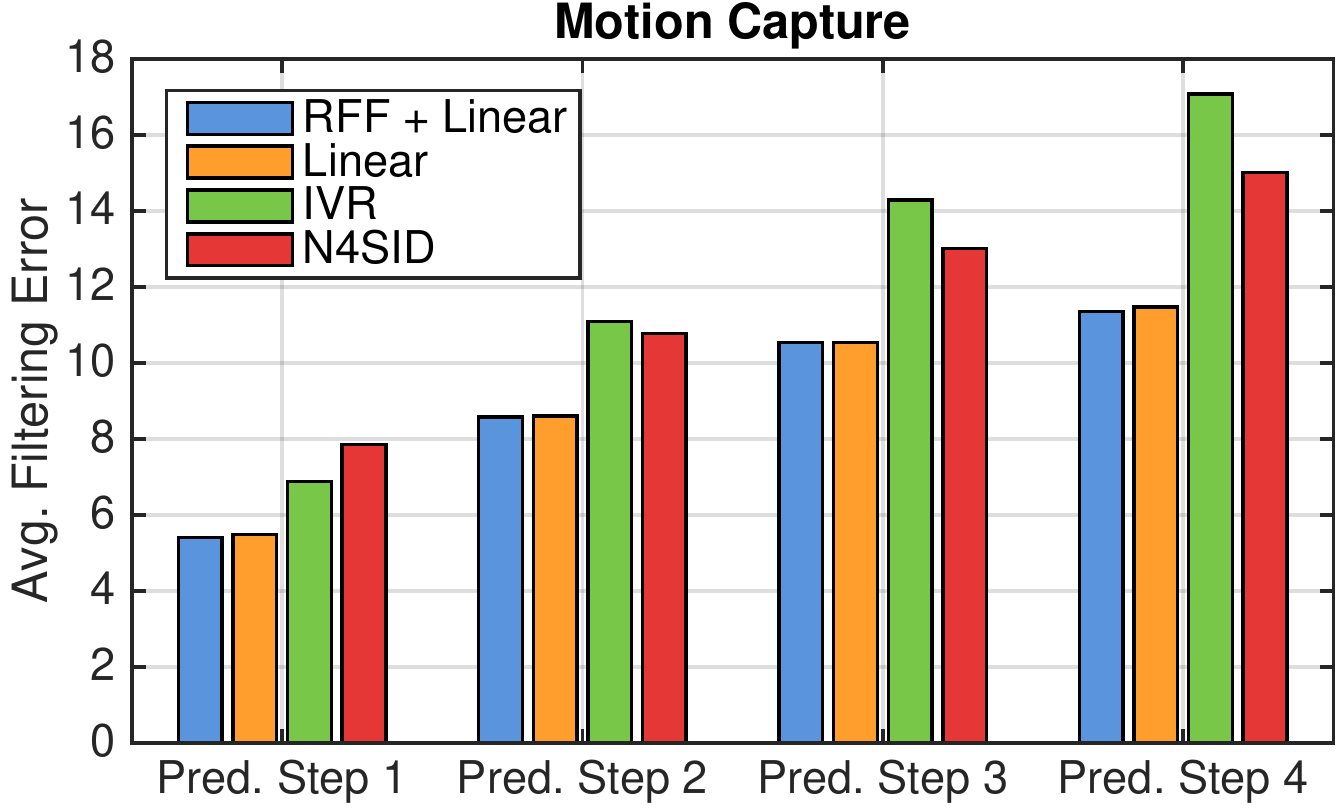}
        \caption{Motion Capture}
    \end{subfigure}
    \vspace{-3mm}
    \caption{Filter error for multiple look ahead steps for the future predictions shown for a few of the datasets. We see across datasets that the performance of both IVR and N4SID are significantly worse than using \pbim with either linear or random Fourier feature $+$ linear learner. For some datasets, the nonlinearity of the random Fourier features helps to improve the performance.\vspace{-2mm}}
    \label{fig:bars}
    \vspace*{-0.1in}
\end{figure*}

Since the data is collected from the stationary Kalman filter of the $2$-observable LDS, we set $k=2$ and use $\phi_1 (f_t)= [x_t,x_{t+1}]$. Note that the 4-dimensional predictive state  $\mathbb{E}[\phi_1(f_t) | h_t]$ will represent the exact conditional distribution of observations $(x_t,x_{t+1})$ and therefore is equivalent to $P(s_t|h_{t-1})$ (see the detailed case study for LDS in Appendix). With linear ridge regression, we test \pbim with forward training, \pbim with DAgger, and AR models (AR-k) with different lengths ($k$ steps of past observations) of history on this dataset. For each method, we compare the average filtering error $e$ to $e_{F}$ which is computed by using the underlying linear filter $F$ of the LDS.


Fig.~\ref{fig:forward_training_convergence} shows the convergence trends of PSIM with DAgger, PSIM with Forward Training, and AR as the number of training trajectories $N$ increases. The prediction error for AR with $k=5,10,20$ is too big to fit into the plot. \pbim with DAgger performs much better with few training data while Forward Training eventually slightly surpasses DAgger with sufficient data. The AR-k models need long histories to perform well given data gnereated by latent state space models, even for this 2-observable LDS. Note AR-35 performs regression in a 70-dimensional feature space (35 past observations), while \pbim only uses 6-d features (4-d predictive state + 2-d current observation). This shows that \emph{predictive state} is a more compact representation of the history and can reduce the complexity of learning problem.  



\vspace{-5pt}
\subsection{Real Dynamical Systems}
We consider the following three real dynamical systems: (1) \emph{Robot Drill Assembly}: the dataset consists of 96 sensor telemetry traces, each of length 350, from a robotic manipulator assembling the battery pack on a power drill. The 13 dimensional noisy observations consist of the robot arm's 7 joint torques as well as the the 3D force and torque vectors. 
Note the fixed higher level control policy for the drill assembly task is not given in the observations and must be learned as part of the dynamics; (2) \emph{Human Motion Capture}:
the dataset consists of 48 skeletal tracks of 300 timesteps each from a Vicon motion capture system from three human subjects performing walking actions. The observations consist of the 3D positions of the various skeletal parts (e.g. upperback, thorax, clavicle, etc.); 
(3) \emph{Video Textures}: the datasets consists of one video of \emph{flag} waving and the other one of waves on a \emph{beach}. 

For these dynamical systems, we do not test \pbim with Forward Training since our benchmarks have a large number of time steps per trajectory. 
Throughout the experiments, we set $k = 5$ for all datasets except for video textures, where we set $k = 3$. For each dataset, we randomly pick a small number of trajectories as a validation set for parameter tuning (e.g., ridge, rank for N4SID and IVR, band width for RFF). 
We partition the whole dataset into ten folds, train all algorithms on 9 folds and test on 1 fold. For the feature function $\phi_1$, the average one-step filtering errors and its standard deviations across ten folds are shown in Tab.~\ref{tab:mean_std}. Our approaches outperforms the two baselines across all datasets. Since the datasets are generated from complex dynamics, \pbim with RFF exhibits better performance than \pbim with Linear. This experimentally supports our theorems suggesting that with powerful regressors, \pbim could perform better. We implement \pbim with back-propagation using Theano with several training approaches: gradient descent with step decay, RMSProp \cite{tieleman2012lecture} and AdaDelta \cite{zeiler2012adadelta} (see Appendix.~\ref{sec:extral_exp}). With random initialization, back-propagation does not achieve comparable performance, except on the flag video, due to local optimality.
We observe marginal improvement by using back-propogation to refine the solution from DAgger.  
This shows \pbim with DAgger finds good models by itself (details in Appendix.~\ref{sec:extral_exp}). We also compare these approaches for multi-step look ahead (Fig.~\ref{fig:bars}). \pbim consistently outperforms the two baselines. 

To show predictive states with larger $k$ encode more information about latent states, we additionally run \pbim with $k=1$ using $\phi_1$ . \pbim (DAgger) with $k=5$ outperforms $k=1$ by 5\% for robot assembly dataset, 6\% for motion capture, 8\% for flag and 32\% for beach video. Including belief over longer futures into predictive states can thus capture more information and increase the performance. 
 
 For feature function $\phi_2$ and $k=5$, with linear ridge regression, the 1-step filter error achieved by \pbim with DAgger across all datasets are: $2.05 \pm 0.08$ on Robot Drill Assembly, $5.47 \pm 0.42$ on motion capture, $154.02 \pm 9.9$ on beach video, and $1.27e3 \pm 13e1$ on flag video. Comparing to the results shown in the \textbf{PSIM-Linear}$_d$ in column of Table.~\ref{tab:mean_std}, we achieve slightly better performance on all datasets, and noticeably better performance on the beach video texture.

\vspace{-5pt}
\section{Conclusion}
\vspace{-5pt}
We introduced \emph{\PBIMs}, a novel approach to directly learn to filter with latent state space models. Leveraging ideas from PSRs, \pbim reduces the unsupervised learning of latent state space models to a supervised learning setting and guarantees filtering performance for general non-linear models in both the realizable and agnostic settings. 



\vspace{-5pt}
\section*{Acknowledgements}
\vspace{-5pt}
This material is based upon work supported in part by: DARPA ALIAS contract number HR0011-15-C-0027 and National Science Foundation Graduate Research Fellowship Grant No. DGE-1252522. The authors also thank Geoff Gordon for valuable discussions.

\small{
\bibliography{reference}
\bibliographystyle{icml2016}
}

\input{appendix}

\end{document}

%% file: appendix.tex
\newpage
\appendix
\onecolumn

\section{Proof of Theorem.~\ref{them:consistent}}
\begin{proof}
We prove the theorem by induction. We start from $t=1$. Under the assumption of infinite many training trajectories, $\hat{m}_1$ is exactly equal to $m_1$, which is $\mathbb{E}_{\tau}(\phi(f_1))$ (no observations yet, conditioning on nothing).  

Now let us assume at time step $t$, we have all computed $\hat{m}_{j}^{\tau}$ equals to $m_{j}^{\tau}$ for $1\leq j\leq t$ on any trajectory $\tau$.  Under the assumption of infinite training trajectories, minimizing the empirical risk over $D_t$ is equivalent to minimizing the true risk $\mathbb{E}_{\tau} [d(F(m_t^{\tau}, x_t^{\tau}), f_{t+1}^{\tau})]$.
Since we use sufficient features for distribution $P(f_t|h_{t-1})$ and we assume the system is $k$-observable, there exists a underlying deterministic map, which we denote as $F_t^*$ here, that maps $m_t^{\tau}$ and $x_t^{\tau}$ to $m_{t+1}^{\tau}$ (Eq.~\ref{eq:message_pass_HMM_PSR} represents $F^*_t$).
Without loss of generality, for any $\tau$, conditioned on the history $h_t^{\tau}$, we have that for a noisy observation $f_t^{\tau}$:
\begin{align}
\phi(f^{\tau}_{t+1}) | h^{\tau}_t& = 
\mathbb{E}[\phi(f^{\tau}_{t+1})|h^{\tau}_t] + \epsilon \\
& = m_{t+1}^{\tau} + \epsilon \\
& = F_t^*(m_t^{\tau}, x_t^{\tau}) + \epsilon,
\end{align}where $\mathbb{E}[\epsilon] = 0$. Hence we have that $F^*_t$ is the operator of conditional expectation $\mathbb{E}[\big(\phi(f_{t+1})|h_t\big)|m_t,x_t]$, which exactly computes the predictive state $m_{t+1} = \mathbb{E}[\phi(f_{t+1}^{\tau})|h_t^{\tau}]$, given $m_t^{\tau}$ and $x_t^{\tau}$ on any trajectory $\tau$.

Since the loss $d$ is a squared loss (or any other loss that can be represented by Bregman divergence), the minimizer of the true risk will be the operator of conditional expectation $\mathbb{E}[\big(\phi(f_{t+1})|h_t\big)|m_t,x_t]$. Since it is equal to $F^*$ and we have $F^*\in \mathcal{F}$ due to the realizable assumption, the risk minimization at step $t$ exactly finds $F^*_t$. Using $\hat{m}_t^{\tau}$ (equals to $m_t^{\tau}$ based on the induction assumption for step $t$), and $x_t^{\tau}$, the risk minimizer $F^*$ then computes the exact $m_{t+1}^{\tau}$ for time step $t+1$. 
Hence by the induction hypothesis, we prove the theorem.
\end{proof}

\section{Proof of Theorem.~\ref{them:forward_train_infinite}}
Under the assumption of infinitely many training trajectories, we can represent the objective as follows:
\begin{align}
\mathbb{E}_{\tau\sim\mathcal{D}}\frac{1}{T}\sum_{t=1}^T d(F_t(\hat{m}_t^{\tau},x_t^{\tau}), f_{t+1}^{\tau}) = \frac{1}{T}\sum_{t=1}^T \mathbb{E}_{(z,f)\sim \omega_t}\big[ d(F_t(z), f)\big]
\end{align}
Note that each $F_t$ is trained by minimizing the risk:
\begin{align}
F_t = \arg\min_{F\sim\mathcal{F}} \mathbb{E}_{(z,f)\sim\omega_{t}}\big[d(F(z),f)\big].
\end{align} Since we define $\epsilon_t = \min_{F\sim\mathcal{F}}\mathbb{E}_{(z,f)\sim \omega_t}\big[d(F(z), f) \big]$, we have:
\begin{align}
    &\mathbb{E}_{\tau\sim\mathcal{D}}\frac{1}{T}\sum_{t=1}^T d(F_t(\hat{m}_t^{\tau},x_t^{\tau}), f_{t+1}^{\tau}) = \frac{1}{T}\sum_{t=1}^T \mathbb{E}_{(z,f)\sim \omega_t}\big[ d(F_t(z), f)\big] \leq \frac{1}{T}\sum_{t} \epsilon_t.
\end{align} Defining $\epsilon_{max} = \max_t\{\epsilon_t\}$, we prove the theorem.

\section{Proof of Theorem.~\ref{them:finite_sample_analysis_forward_training}}
\label{sec:proofs}

\begin{proof}
Without loss of generality, let us assume the loss $d(F(z),f)\in [0,1]$. 
To derive generalization bound using Rademacher complexity, we assume that $\|F(z)\|_2$ and $\|f\|_2$ are bounded for any $z,f, F\in\mathcal{F}$, which makes sure that $d(F(z), f)$ will be Lipschitz continuous with respect to the first term $F(z)$\footnote{Note that in fact for the squared loss, $d$ is $1$-smooth with respect to its first item. In fact we can remove the boundness assumption here by utilizing the existing Rademacher complexity analysis for smooth loss functions \citep{srebro2010optimistic}. }.  

Given $M$ samples, we further assume that we split $M$ samples into $T$ disjoint sets $S_1, ..., S_T$, one for each training process of $F_i$, for $1\leq i\leq T$.
The above assumption promises that the data $S_t$ for training each filter $F_t$ is i.i.d. Note that each $S_i$ now contains $M/T$ i.i.d trajectories. 


Since we assume that at time step $t$, we use $S_t$ (rolling out $F_1, ..., F_{t-1}$ on trajectories in $S_t$) for training $F_t$, we can essentially treat each training step independently: when learning $F_t$, the training data $z, f$ are sampled from $\omega_t$ and are i.i.d.

Now let us consider time step $t$. With the learned $F_1, ..., F_{t-1}$, we roll out them on the trajectories in $S_t$ to get $\frac{M}{T}$ i.i.d samples of $(z, f)\sim \omega_t$. Hence, training $F_t$ on these $\frac{M}{T}$ i.i.d samples becomes  classic empirical risk minimization problem. 
Let us define loss class as $\mathcal{L} = \{ l_F: (z, f) \rightarrow d(F(z), f): F\in\mathcal{F}\}$, which is determined by $\mathcal{F}$ and $d$. Without loss of generality, we assume $l(z,f) \in [0,1], \forall l\in\mathcal{L}$. 
Using the uniform bound from Rademacher theorem \cite{mohri2012foundations}, we have for any $F\in\mathcal{F}$, with probability at least $1-\delta'$:
\begin{align}
&\mathbb{E}_{z,f\sim \omega_t} [d(F(z), f)] - \frac{T}{M}\sum_{i} d(F(z^i), f^i)| \\
&\leq 2\mathcal{R}_t(\mathcal{L}) + \sqrt{\frac{T\ln(1/\delta')}{2M}},
\end{align} where $\mathcal{R}_t(\mathcal{L})$ is Rademacher complexity of the loss class $\mathcal{L}$ with respect to distribution $\omega_t$. Since we have $F_t$ is the empirical risk minimizer, for any $F_t^*\in\mathcal{F}$, we have with probability at least $1-\delta'$:
\begin{align}
&\mathbb{E}_{z,f\sim \omega_t} [d(F_t(z), f)] \leq  
\mathbb{E}_{z,f\sim \omega_t}[d(F_t^*(z^i), f^i)] + 4\mathcal{R}_t(\mathcal{L}) + 2\sqrt{\frac{T\ln(1/\delta')}{2M}}.
\label{eq:single_step}
\end{align}

Now let us combine all time steps together.  For any $F_t^*\in\mathcal{F}$, $\forall t$, with probability at least $(1-\delta')^T$, we have:
\begin{align}
\label{eq:rademacher_with_L}
&\mathbb{E}_{\tau\sim\mathcal{D}_{\tau}}\Big[  \frac{1}{T}\sum_{t=1}^Td(F_t(\hat{m}_t^{\tau}, x_t^{\tau}), f_{t+1}^{\tau})  \Big]   = \frac{1}{T}\sum_{t=1}^T\mathbb{E}_{z,f\sim d_{t}}\big[d(F_t(z), f) \big]  \nonumber \\
&\leq \frac{1}{T}\sum_{t=1}^T\mathbb{E}_{z,f\sim \omega_t}[d(F_t^*(z), f)] + 4\bar{\mathcal{R}}(\mathcal{L}) + 2\sqrt{\frac{T\ln(1/\delta')}{2M}}  \nonumber
\\& = \mathbb{E}_{\tau\sim \mathcal{D}_{\tau}}\big[ \frac{1}{T}\sum_{t=1}^T d(F_t^*(\hat{m}_t^{\tau}, x_t^{\tau}), f_{t+1}^{\tau})   \big] +4\bar{\mathcal{R}}(\mathcal{L})  + 2\sqrt{\frac{T\ln(1/\delta')}{2M}},
\end{align} where $\bar{\mathcal{R}}(\mathcal{L}) = (1/T)\sum_{t=1}^T \mathcal{R}_t(\mathcal{L})$ is the average Rademacher complexity.  Inequality.~\ref{eq:rademacher_with_L} is derived from the fact the event that the above inequality holds can be implied by the event that Inequality.~\ref{eq:single_step} holds for every time step $t$ ($1\leq t\leq T$)  independently. The probability of Inequality.~\ref{eq:single_step} holds for all $t$ is at least $(1-\delta')^T$.

Note that in our setting $d(F(z), f) = \|F(z) - f\|_2^2$, and under our assumptions that $\|F(z)\|_2$ and $\|f\|_2$ are bounded for any $z, f, F\in\mathcal{F}$, $d(F(z), f)$ is Lipschitz continuous with respect to its first item with Lipschitz constant equal to $\nu$, which is $\sup_{F,z,f} 2\|F(z) - f\|_2$. Hence, from the composition property of Rademacher number \cite{mohri2012foundations}, we have:
\begin{align}
\mathcal{R}_t({\mathcal{L}} )\leq \nu\mathcal{R}_t(\mathcal{F}), \;\;\;\; \forall t.
\label{eq:lip_rad}
\end{align} 

It is easy to verify that for $T\geq 1$, $\delta' \in (0,1)$, we have $(1-\delta')^T \geq 1 - T\delta'$.  Let $1 - T\delta' = 1 - \delta$, and solve for $\delta'$, we get $\delta' = \delta / T$. Substitute Eq.~\ref{eq:lip_rad} and $\delta' = \delta / T$ into Eq.~\ref{eq:rademacher_with_L}, we prove the theorem.
\end{proof}

Note that the above theorem shows that for fixed number training examples, the generalization error increase as $\tilde{O}(\sqrt{T})$ (sublinear with respect to $T$).


\section{Case Study: Stationary Kalman Filter}
\label{sec:case_study}
To better illustrate \pbim, we consider a special dynamical system in this section. More specifically, we focus on the stationary Kalman filter \citep{Boots2012_Thesis,NIPS2015_5673} \footnote{For a well behaved system, the filter will become stationary (Kalman gain converges) after running for some period of time. Our definition here is slightly different from the classic Kalman filter: we focus on filtering from $P(s_t |h_{t-1})$ (without conditioning on the  observation $x_t$ generated from $s_t$) to $P(s_{t+1}|h_t)$, while traditional Kalman filter usually filters from $P(s_t|h_t)$ to $P(s_{t+1}|h_{t+1})$.}: 
\begin{align}
\label{eq:linear_sys_appendix}
s_{t+1} = A s_{t} + \epsilon_s, \;\; \epsilon_s \sim \mathcal{N}(0, Q),\nonumber \\
x_{t} = Cs_{t} + \epsilon_x, \;\; \epsilon_x \sim \mathcal{N}(0, R). 
\end{align} 
As we will show, the Stationary Kalman Filter allows us to explicitly represent the predictive states (sufficient statistics of the distributions of future observations are simple). We will also show that we can explicitly construct a bijective map between the predictive state space and the latent state space, which further  enables us to explicitly construct the predictive state filter. We will show that the predictive state filter is closely related to the original filter in the latent state space.

The $k$-observable assumption here essentially means that the observability matrix:
$\mathcal{O} = \begin{bmatrix}
C & CA & CA^2 &... & CA^{k-1}
\end{bmatrix}^\top$ is full (column) rank. 
Now let us define $P(s_t |h_{t-1}) = \mathcal{N}(\hat{s}_t, \Sigma_{s})$, and $P(f_t | h_{t-1}) = \mathcal{N}(\hat{f}_t, \Sigma_{f})$. Note that $\Sigma_{s}$ is a constant for a stationary Kalman filter (the Kalman gain is converged). Since $\Sigma_{f}$ is purely determined by $\Sigma_s$, $A$, $C$, $R$, $Q$, it is also a constant. It is clear now that $\hat{f}_t = \mathcal{O}\hat{s}_t$. When the Kalman filter becomes stationary, it is enough to keep tracking $\hat{s}_t$.  Note that here, given $\hat{s}_t$, we can compute $\hat{f}_t$; and given $\hat{f}_t$, we can reveal $\hat{s}_t$ as $\mathcal{O}^{\dagger}\hat{f}_t$, where $\mathcal{O}^{\dagger}$ is the pseudo-inverse of $\mathcal{O}$. This map is bijective since $\mathcal{O}$ is full column rank due to the $k$-observability. 


Now let us take a look at the update of the stationary Kalman filter: 
\begin{align}
&\hat{s}_{t+1} = A\hat{s}_t - A\Sigma_{s}C^T(C\Sigma_{s}C^T + R)^{-1}(C\hat{s}_t - x_t) = A\hat{s}_t - L (C\hat{s}_t - x_t),
\label{eq:skf_latent}
\end{align} where we define $L = A\Sigma_s C^T(C\Sigma_{s}C^T+R)^{-1}$. Here due to the stationary assumption, $\Sigma_s$ keeps constant across time steps. 
Multiple $\mathcal{O}$ on both sides and plug in $\mathcal{O}^{\dagger}\mathcal{O}$, which is an identity,  at proper positions, we have: 
\begin{align}
&\hat{f}_{t+1} =  \mathcal{O}\hat{s}_{t+1} = \mathcal{O}A (\mathcal{O}^{\dagger}\mathcal{O}) \hat{s}_t - \mathcal{O}L  (C\mathcal{O}^{\dagger}\mathcal{O}\hat{s}_t - x_t) \nonumber \\
& =\mathcal{O}A\mathcal{O}^{\dagger} \hat{f}_t -\mathcal{O}L (C\mathcal{O}^{\dagger}\hat{f}_{t} - x_t) = \tilde{A}\hat{f}_t - \tilde{L}(\tilde{C}\hat{f}_t - x_t) \\
& = \begin{bmatrix} \tilde{A} - \tilde{L}\tilde{C} & \tilde{L} \end{bmatrix}\begin{bmatrix} \hat{f}_t \\ x_t  \end{bmatrix}, 
\label{eq:skf_final}
\end{align}
where we define $\tilde{A} = \mathcal{O}A\mathcal{O}^{\dagger}$, $\tilde{C} = C\mathcal{O}^{\dagger}$ and $\tilde{L} = \mathcal{O}L$. 
The above equation represents the \emph{stationary} filter update step in predictive state space. Note that the \emph{deterministic} map from $(\hat{f}_t, \Sigma_{f})$ and $x_t$ to $(\hat{f}_{t+1}, \Sigma_{f})$ is a linear map ($F$ defined in Sec.~\ref{sec:psim} is a linear function with respect to $\hat{f}_t$ and $x_t$). The filter update in predictive state space is very similar to the filter update in the original latent state space except that predictive state filter uses operators ($\tilde{A}, \tilde{C}, \tilde{Q}$) that are linear transformations of the original operators ($A, C, Q$).   

We can do similar linear algebra operations (e.g., multiply $\mathcal{O}$ and plug in $\mathcal{O}^{\dagger}\mathcal{O}$ in proper positions) to recover the stationary filter in the original latent state space from the stationary predictive state filter. The above analysis leads to the following proposition:
\begin{proposition}
\label{prop:LDS_equal}
For a linear dynamical system with $k$-observability, there exists a filter in predictive state space (Eq.~\ref{eq:skf_final}) that is equivalent to the stationary Kalman filter in the original latent state space (Eq.~\ref{eq:skf_latent}).
\end{proposition}

We just showed a concrete bijective map between the filter with predictive states and the filter with the original latent states by utilizing the observability matrix $\mathcal{O}$. Though we cannot explicitly construct the bijective map unless we know the parameters of the LDS (A,B,C,Q,R), we can see that learning the linear filter shown in Eq.~\ref{eq:skf_final} is equivalent to learning the original linear filter in Eq.~\ref{eq:skf_latent} in a sense that the predictive beliefs filtered from Eq.~\ref{eq:skf_final} encodes as much information as the beliefs filtered from Eq.~\ref{eq:skf_latent} due to the existence of a bijective map between predictive states and the beliefs for latent states.


\subsection{Collection of Synthetic Data}
 \label{sec:synthetic_data_collect}
We created a linear dynamical system with $A\in\mathbb{R}^{3\times 3}$, $C\in\mathbb{R}^{2\times 3}$, $Q\in\mathbb{R}^{3\times 3}$, $R\in\mathbb{R}^{2\times 2}$. The matrix $A$ is full rank and its largest eigenvalue is less than $1$.  The LDS is $2$-observable.  We computed the constance covariance matrix $\Sigma_{s}$, which is a fixed point of the covariance update step in the Kalman filter.  The initial distribution of $s_0$ is set to $\mathcal{N}(\texttt{1}, \Sigma_s)$.  We then randomly sampled 50000 observation trajectories from the LDS.  We use half of the trajectories for training and the left half for testing.

\section{Additional Experiments}
\label{sec:extral_exp}

With linear regression as the underlying filter model: $\hat{m}_{t+1} = W [\hat{m}^T_t, x^T_t]^T$, where $W$ is a 2-d matrix, we compare \pbim with back-propagation using the solutions from DAgger as initialization to \pbim with DAgger, and \pbim with back-propagation with random initialization. We implemented \pbim with Back-propagation in Theano \cite{Bastien-Theano-2012}. For random initialization, we uniformly sample non-zero small matrices to avoid gradient blowing up. For training, we use mini-batch gradient descent where each trajectory is treated as a batch. We tested several different gradient descent approaches: regular gradient descent with step decay, AdaGrad~\cite{duchi2011adaptive}, AdaDelta~\cite{zeiler2012adadelta}, RMSProp \cite{tieleman2012lecture}. We report the best performance from the above approaches.  
When using the solutions from \pbim with DAgger as an initialization for back-propagation, we use the same setup. We empirically find that RMSProp works best across all our datasets for the inference machine framework, while regular gradient descent generally performs the worst. 

\begin{table*}[h]
\begin{center}
    \begin{tabular}{| l | l | l | l | }
    \hline
     & \textbf{\pbim-Linear} (DAgger) &\textbf{\pbim-Linear} (Bp)  & \textbf{\pbim-Linear} (DAgger + Bp) \\ \hline
    \textbf{Robot Drill Assembly} & 2.15 &  2.54 & \textbf{2.09} \\ \hline 
    \textbf{Motion Capture} & 5.75 & 9.94 &\textbf{5.66} \\ \hline
    \textbf{Beach Video Texture} &  164.23 & 268.73  & \textbf{164.08} \\ \hline
    \end{tabular}
\end{center}
\vspace{-10pt}
\caption{Comparison between \pbim with DAgger, \pbim with back-propagation using random initialization, and \pbim with back-propagation using DAgger as initialization with ridge linear regression.
}
\label{tab:extra_exp}
\end{table*}

Tab.~\ref{tab:extra_exp} shows the results of using different training methods with ridge linear regression as the underlying model. 

Additionally, we test back-propagation for \pbim with Kernel Ridge regression as the underlying model: $\hat{m}_{t+1} = W\eta(\hat{m}_t, x_t)$, where $\eta$ is a pre-defined, deterministic feature function that maps $(\hat{m}_t,x_t)$ to a reproducing kernel Hilbert space approximated with Random Fourier Features (RFF). Essentially, we lift the inputs $(\hat{m}_t, x_t)$ into a much richer feature space (a scaled, and transition invariant feature space) before feeding it to the next module. The results are shown in Table.~\ref{tab:extra_exp_RFF}. As we can see, with RFF, back-propagation achieves better performance than back-propagation with simple linear regression (\pbim-Linear (Bp)). This is expected since using RFF potentially captures the non-linearity in the underlying dynamical systems. On the other hand, \pbim with DAgger achieves better results than back-propagation across all the datasets. This result is consistent with the one from \pbim with ridge linear regression.

\begin{table*}[h]
\begin{center}
    \begin{tabular}{| l | l | l | l | l | }
    \hline
     &\textbf{\pbim-RFF} (Bp)  & \textbf{\pbim-RFF} (DAgger) & \textbf{RNN} \\ \hline
    \textbf{Robot Drill Assembly} & 2.54 &  \textbf{1.80} & 1.99 \\ \hline 
    \textbf{Motion Capture} & 9.26 & \textbf{5.41} & 9.6 \\ \hline
    \textbf{Beach Video Texture} &  202.10 & \textbf{130.53} & 346.0  \\ \hline
    \end{tabular}
\end{center}
\vspace{-10pt}
\caption{Comparison between \pbim with DAgger, \pbim with back-propagation using random initialization with kernel ridge linear regression, and Recurrent Neural Network. For RNN, we use 100 hidden states for Robot Drill Assembly, 200 hidden states for motion capture, and 2500 hidden states for Beach Video Texture. 
}
\label{tab:extra_exp_RFF}
\end{table*}

Overall, several interesting observations are: (1) back-propagation with random initialization achieves reasonable performance (e.g., good performance on flag video compared to baselines), but worse than the performance of \pbim with DAgger. \pbim back-propagation is likely stuck at locally optimal solutions in some of our datasets; (2) \pbim with DAgger and Back-propagation can be symbiotically beneficial: using back-propagation to refine the solutions from \pbim with DAgger improves the performance. Though the improvement seems not significant over the 400 epochs we ran, we do observe that running more epochs continues to improve the results; (3) this actually shows that \pbim with DAgger itself finds good filters already, which is not surprising because of the strong theoretical guarantees that it has.
